# 1
# DISEÑO DE DISPOSITIVO ROBÓTICO PARA LA REHABILITACIÓN Y DIAGNOSIS DE EXTREMIDADES INFERIORES

*Pedro Araujo Gómez[1], Miguel Díaz Rodríguez[2], Vicente Mata Amela[3]*

En la actualidad, la rehabilitación robótica para extremidades inferiores se encuentra ampliamente desarrollada; sin embargo, los dispositivos utilizados hasta ahora parecieran no tener criterios uniformes para su diseño, pues, por el contrario, cada mecanismo desarrollado suele presentarse como si no tomara en cuenta los criterios empleados en diseños anteriores. Por otra parte, la diagnosis de extremidades inferiores a partir de dispositivos robóticos ha sido poco estudiada. Este capítulo presenta una guía para el diseño de dispositivos robóticos en la rehabilitación y diagnosis de extremidades inferiores, tomando en cuenta la movilidad de la pierna humana y las técnicas utilizadas por los fisioterapeutas en la ejecución de ejercicios de rehabilitación y pruebas de diagnosis, así como las recomendaciones dadas por diversos autores, entre otros aspectos.

La guía propuesta se ejemplifica mediante un caso de estudio basado en un robot paralelo RPU+3UPS capaz de hacer movimientos que se aplican durante los procesos de rehabilitación y diagnosis. La propuesta presenta ventajas con respecto a algunos dispositivos existentes, como su capacidad de carga que puede soportar, y también permite restringir el movimiento en las direcciones exigidas por la rehabilitación y la diagnosis.

[1] Departamento de Ciencias Aplicadas y Humanísticas, Facultad de Ingeniería, Universidad de Los Andes, Mérida, Venezuela. {pfaraujo@ula.ve}

[2] Laboratorio de Mecatrónica y Robótica, Facultad de Ingeniería, Universidad de Los Andes, Mérida, Venezuela. {dmiguel@ula.ve}

[3] Centro de Investigación en Ingeniería Mecánica, Universitat Politècnica de València, Valencia, España. {vmata@mcm.upv.es}

Palabras clave: Rehabilitación, diagnosis robótica, extremidades inferiores, robot paralelo.

## INTRODUCCIÓN

La robótica ha tenido un desarrollo significativo en las últimas décadas, entre otros aspectos debido a la evolución que han tenido diversas áreas vinculadas a la robótica como, por ejemplo, la informática. Este desarrollo ha incidido tanto en la mejora de los dispositivos robóticos como en la creación de nuevos campos de aplicación, y principalmente se ha extendido a campos relacionados con los servicios y necesidades humanas como agricultura, sistemas de seguridad y aplicaciones médicas.

La rehabilitación robótica se presenta como uno de los campos de mayor interés en la actualidad, sirviendo de asistencia al trabajo arduo de los fisioterapeutas, además de que logra una mejor coordinación para los ejercicios de rehabilitación y mayor precisión en el diagnóstico de lesiones y la medición de la evolución de los pacientes. La rehabilitación y diagnosis de las extremidades inferiores es muy frecuente debido a la gran cantidad de accidentes a los que están expuestas estas extremidades, de hecho, en el campo de los deportes suelen presentarse muy a menudo.

El propósito de este capítulo es presentar una guía para el diseño de dispositivos en la rehabilitación y diagnosis de extremidades inferiores. Se toma como referencia la movilidad de las piernas, el movimiento de la extremidad durante una sesión de rehabilitación y en una prueba diagnóstica, así como también las recomendaciones de diferentes autores.

En la primera parte de este capítulo se encuentra una clasificación de estos mecanismos según ciertos criterios y luego se presenta la revisión de dispositivos existentes tanto para la rehabilitación como para la diagnosis. Seguidamente se muestra una guía para el diseño de dispositivos de rehabilitación y diagnosis de extremidades inferiores; en esta parte se ofrecen recomendaciones de los autores para conseguir un diseño adecuado, así como otras recomendaciones que otros autores han reseñado.

Finalmente se presenta una propuesta de un dispositivo en particular seguido de las conclusiones obtenidas de esta investigación.

## CLASIFICACIÓN DE LOS DISPOSITIVOS DE REHABILITACIÓN Y DIAGNOSIS ROBÓTICA

El campo de la rehabilitación robótica está generalmente dividido en las categorías de robots de terapia y robots de asistencia (van der Loos y Reinkensmeyer, 2008), siendo los robots de terapia los utilizados para las personas que requieren recuperar la movilidad y fuerza de una extremidad lesionada. Ese será, pues, el tema de estudio en este capítulo. Los robots de asistencia cumplen con apoyar a las personas para hacer tareas diarias que no son capaces de llevar a cabo debido a alguna discapacidad presente.

En la actualidad hay gran cantidad de dispositivos para la rehabilitación de extremidades inferiores, las cuales se diferencian entre sí por la fase del tratamiento a la que este destinada (Díaz, Gil y Sánchez, 2011) y que presentan una clasificación de los dispositivos según el principio de rehabilitación al que estén destinados; así, definen 5 grupos, caminadoras en cinta, caminadoras en base para el pie, caminadoras en superficie, caminadoras estacionarias y entrenador para tobillo, y órtesis activas para el pie (*treadmill gait trainers, foot-plate-based gait trainers, overground gaittrainers, stationary gait and ankle trainers, y active foot orthoses*). Alcocer y col., 2012, describen una clasificación de dispositivos de rehabilitación basada en la complejidad de su mecanismo, encontrándose tres grupos, dispositivos de baja complejidad, de complejidad intermedia y de alta complejidad.

Por otro lado, considerando las fases de tratamiento de rehabilitación habrá dispositivos para la fase inicial del tratamiento en la cual la persona lesionada tiene muy poca o nula capacidad de movimiento. Luego están los dispositivos de fase de rehabilitación, en los que el paciente comienza a recuperar la fuerza y movilidad de la extremidad. Por último, vemos dispositivos para la fase funcional que buscan completar la movilidad de la extremidad. Este capítulo está enmarcado en el desarrollo de un dispositivo destinado para la fase intermedia de la rehabilitación, por la que se recupera la fuerza y la movilidad de la extremidad. Sin embargo, este dispositivo puede aplicarse a varias fases del tratamiento de rehabilitación.

Los dispositivos de diagnosis suelen clasificarse con base en la prueba a la que estén destinados, entre las que destaca el test de Lachman y el test de desplazamiento de pivote. A continuación hacemos una revisión de los principales dispositivos desarrollados para la rehabilitación y para la diagnosis.

Dispositivos actuales para la rehabilitación

Existe una gran variedad de dispositivos para la rehabilitación de extremidades inferiores, algunos de ellos comerciales y otros en fase de investigación. A pesar de que diversos autores han desarrollado criterios para clasificarlos, en ocasiones, algunos de ellos pueden estar comprendidos en varios grupos.

Los dispositivos comerciales más desarrollados son los que principalmente asisten el proceso de marcha, así como también los de tipo estacionario, y por lo general presentan costos elevados y principalmente enfocados hacia su implementación en centros de salud. Entre los más conocidos se encuentra el Lokomat, desarrollado por Hocoma AG, Volketswil, Suiza, el cual consiste de una órtesis robótica para asistir la marcha, tal como se observa en la figura 1. Es este un sistema avanzado para el soporte del cuerpo combinado con una cinta caminadora (Díaz, Gil y Sánchez, 2011). Similares al Lokomat son el MotionMaker TM, el Walk Trainer, POGO, AutoAmbultor (HealthSouth Coopertion), LokoHelp (LokoHelp Group).

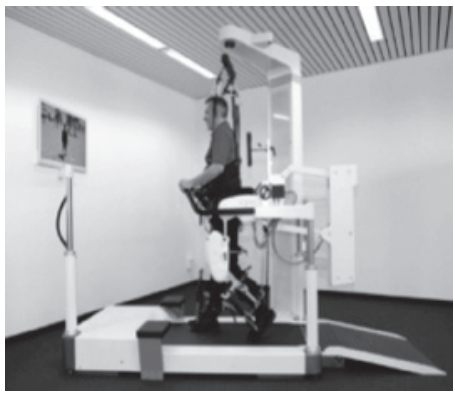

Figura 1. Lokomat de Hocoma AG (Díaz, Gil y Sánchez, 2011)

El MotionMaker TM es un dispositivo estacionario para pacientes discapacitados y hemipléjicos; consta de un sistema de electroestimulación para activar la movilización de las extremidades inferiores en tiempo real (Schmitt y Métrailler, 2004) y fue desarrollado por la Escuela Politécnica Federal de Lausana. Este dispositivo contribuye tanto a la diagnosis como a la recuperación de las funciones durante el proceso de rehabilitación.

El WalkTrainer está también destinado a pacientes con parálisis, y a diferencia del MotionMaker, el paciente se encuentra en posición de pie. El WalkTrainer asemeja el movimiento natural de la extremidad inferior, y esto incide en la motivación del paciente. El dispositivo está compuesto de distintos elementos, entre los que se encuentra un marco para el caminado, una órtesis pélvica, un sistema para el soporte del cuerpo, dos órtesis para las piernas y un sistema para la electroestimulación (Stauffer y col., 2009).

La italiana Easytech, S.R.L. ha desarrollado dos mecanismos: el Genu 3 y el Primadoc. Ambos fueron diseñados para el uso diario y permiten al músculo ejercitarse a una velocidad constante a lo largo del rango de movimiento de la articulación, generando una fuerza de resistencia que es función de la fuerza resistiva generada por el paciente. Esto permite que el músculo siempre desarrolle altas tensiones, lo que incide en una máxima estimulación de las fibras (www.easytechitalia.com). En la figura 2 se puede observar el Genu 3.

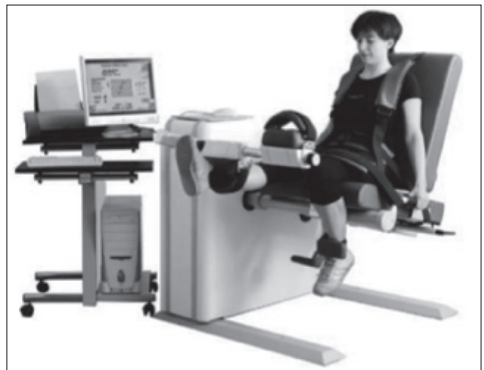

Figura 2. Genu 3 de la Easytech, S.R.L. (www.easytechitalia.com)

Yaskawa Electric produjo un dispositivo para la rehabilitación de extremidades inferiores, el TEM LX2, que comenzó a desarrollarse desde hace más de 15 años. Se trata de un dispositivo que permite hacer ejercicios de movimiento pasivo continuo (CPM) y otros que usualmente se aplican por terapeutas. El dispositivo es de tipo serial y sobre él se fija la pierna del paciente a la que se hace el tratamiento de rehabilitación (Sakaki, 1999). Luego de comercializado en el año 2004, no se reportan cambios importantes y en la actualidad no se observa que el dispositivo se ofrezca por parte de la empresa.

En cuanto a dispositivos que aún se encuentran en fase de investigación, algunos con gran desarrollo y otros básicamente en una fase inicial, hay una gran diversidad de dispositivos en cuanto a la fase de rehabilitación a la que están destinados.

Destaca un dispositivo estacionario para la rehabilitación de rodillas desarrollado por Akdogan y Adli (2011). Este trabaja en distintos modos, uno definido como modo de enseñanza, en el que el terapeuta manipula el dispositivo asemejando ciertos ejercicios de rehabilitación y de esta manera el dispositivo recoge la información de fuerzas y posiciones de la terapia. Luego, en el modo terapia, el robot es capaz de controlar al paciente tomando en cuenta las fuerzas de reacción que este pudiera generar. Además, considera los límites que debe tener el movimiento ejecutado durante el proceso de terapia (Fig. 3).

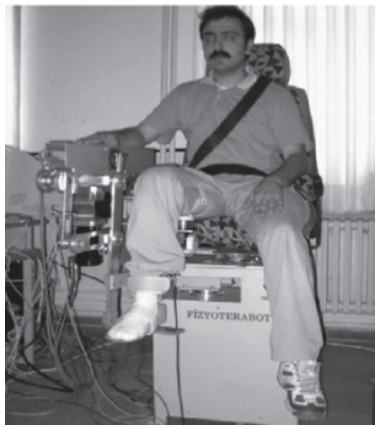

Figura 3. Physiotherabot (Akdogan y Adli 2011)

LOPES es un dispositivo que combina un segmento que permite la traslación de la pelvis libremente con un segmento exoesquelético con tres actuadores rotacionales. Este tiene dos modos de ejecución, definidos como paciente en carga y robot en carga, que permiten tanto que el dispositivo sea controlado por el paciente o que el robot guíe al paciente (Veneman y col., 2007). Este mecanismo se encuentra en la categoría de los *walktrainer*, pero también es de tipo exoesquelético, que se entiende como otra categoría de mecanismos de rehabilitación y es un formato que ha sido utilizado frecuentemente para el desarrollo de mecanismos.

El LAMBDA, desarrollado por Bouri, Le Gall, y Clavel (2009) se puede observar en la Fig. 4, y consiste en un sistema de dos articulaciones traslacionales y una rotacional permitiendo la movilización de la extremidad inferior en el plano sagital, lo que limita en cierta medida la gama de ejercicios del paciente, aunque presenta la ventaja de que también puede ser empleado como equipo de entrenamiento para actividades deportivas. Además, sus autores prevén incluir escenarios de simulación de ejercicios, así como también sensores de fuerza y movimiento para el monitoreo de paciente.

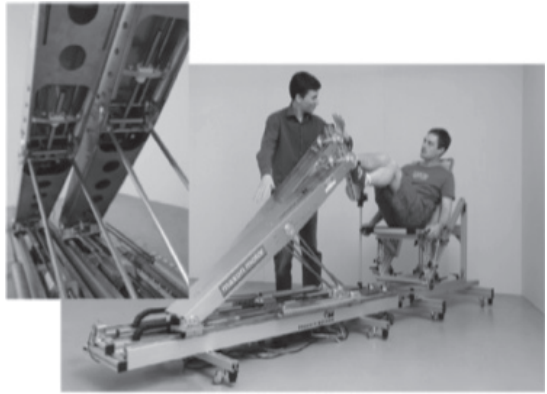

Figura 4. LAMBDA (Bouri, Le Gall, & Clavel, 2009)

Similar al Lambda es el NeXOS (Bradley y col., 2009.), que trabaja en el plano sagital y solo tiene 2 GDL pero se plantea aumentarlo a 4 GDL; además, este proyecto propone que el dispositivo pueda ser teleoperado para su utilización en casa.

Todos los dispositivos descritos hasta ahora coinciden en el hecho de tener una configuración bastante robusta, tanto los comerciales como los que están en fase de investigación. El NeXOS es el más compacto de los dispositivos estacionarios que se han descrito.

Los dispositivos de tipo exoesquelético suelen ser más compactos, como es el caso del SKAFO (Font-Llagunes, Arroyo, Alonso y Vinagre, 2010) (ver Fig. 5), destinado para asistir a la persona durante la fase de balanceo, tanto en la flexión como en la extensión. Además de poseer un sistema antiequino, consta de dos grados de libertad (GDL), y a partir de una serie de sensores, determina en qué fase del caminado se encuentra la persona. El AKROD (Weinberg y col., 2007), diseñado para pacientes con apoplejía, ayuda a corregir la hiperextensión durante la fase de estancia en el caminado y también la poca flexión durante el balanceo. Este dispositivo se basa en un fluido electrorreológico para aplicar un sistema de freno, y no tiene motores que provoquen la actuación del dispositivo. Estos dispositivos (SKAFO, AKROD) solo asisten en cierta medida durante el proceso de caminado del paciente.

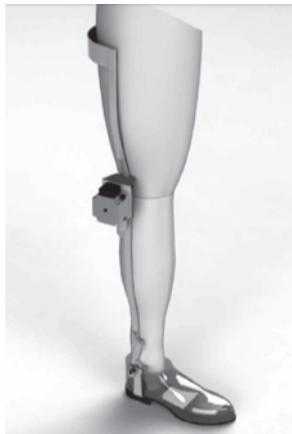

Figura 5. SKAFO (Font-Llagunes y col., 2010)

Finalmente, entre otros dispositivos que se encuentran en fase de investigación se pueden destacar el LEG-100 (Michnik y col, 2012), que resulta ser menos robusto que algunos de los descritos anteriormente.

Valdivia y col. (2013) proponen un dispositivo de cinco GDL que incluye la movilidad de la cadera y de la rodilla y es capaz de planificar trayectorias para simular ejercicios de rehabilitación tanto bidimensional como tridimensional, es decir, que no se limita al plano sagital. Wang y col. (2014) presentan un mecanismo que solo se mueve en el plano sagital, pero que tiene la particularidad de que genera movimientos rotacionales a partir de pares prismáticos.

Una vez observados los dispositivos que en la actualidad existen o que han sido propuestos, se puede aseverar que no existe uniformidad de criterios en el diseño de los dispositivos de rehabilitación para rodillas, pues cada autor presenta ventajas del dispositivo que propone. En general, los mecanismos se encuentran en una fase de investigación medianamente avanzada.

## DISPOSITIVOS ACTUALES PARA LA DIAGNOSIS

El diagnóstico de lesiones de rodillas se fundamenta básicamente en observar el estado en el que se encuentran los ligamentos cruzados anteriores (LCA). Al lesionarse los LCA, dejan de limitar los movimientos de traslación y rotación de la tibia respecto al fémur y, en consecuencia, la inestabilidad en la rodilla es un indicador aceptado de daño en los LCA, así como para determinar la evolución del paciente tras una operación de reconstrucción del ligamento.

Se han desarrollado test específicos para diagnóstico, los dos más reconocidos son el test de Lachman y el test de desplazamiento del pivote (Pivot Shift Test). El de Lachman es el más difundido y se emplea fundamentalmente para estimar la laxitud de los LCA después de una intervención. Este test es estático y diagnostica con base en los desplazamientos relativos entre la tibia y el fémur en el plano sagital, mientras que el test de desplazamiento del pivote pretende reproducir la inestabilidad traslacional y rotacional en la rodilla, aplica una torsión a la tibia y mide esencialmente la rotación. Sin embargo, la rotación de la tibia es difícil de medir de un modo preciso y repetible, y al parecer se ve muy afectada después de la reconstrucción de los LCA. Hay que tener en cuenta que lo que se intenta medir es un movimiento combinado de rotación y traslación.

Entre los dispositivos que se basan en el test de Lachman, uno de los más antiguos es el KT-1000 para medir la laxitud anterior posterior en la rodilla y data de 1985 (ver Fig. 6). Está orientado específicamente a reproducir y cuantificar el test de Lachman, por lo que solo trabaja en el plano sagital. Es el más referenciado y comentado, aunque no siempre de un modo positivo (Van Thiel y Bach, 2010).

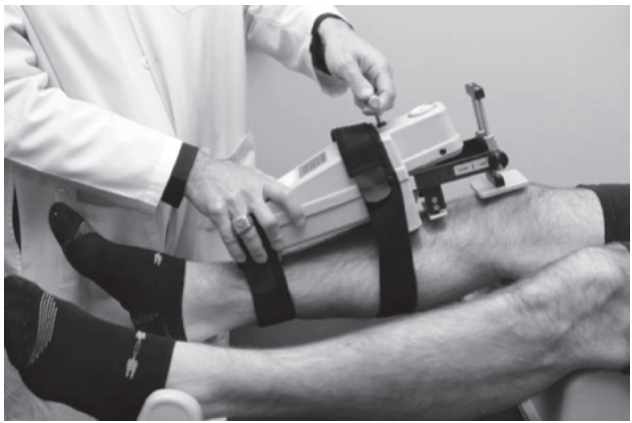

Figura 6. KT-1000.( http://drrobertlaprademd.com/kt-1000-testing-for-acl-tear/)

A diferencia del KT-1000, el Acufex Knee Signature System (KSS) permite medir la laxitud en diferentes planos. Consiste en un electrogoniómetro con cuatro GDL para la medida de la traslación tibiofemoral (ver Fig. 7). Una vez amarrado a la pierna del sujeto se puede someter la pierna a fuerzas en varias direcciones y el aparato efectúa mediciones de traslaciones anteroposteriores, rotaciones varus-valgus y flexión. Una ventaja del aparato es que el sujeto lo puede llevar puesto durante su actividad más o menos normal. El Acufex KSS ha sido descontinuado y el CA-4000 Electrogoniometer (OS Inc, Hayward CA) es el producto actual con tecnología similar (van Thiel y Banch, 2010).

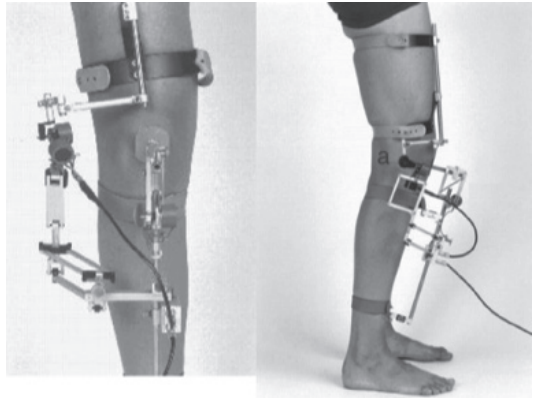

Figura 7. CA-4000 Electrogoniometer (Van Thiel y Banch, 2010)

En cuanto a la medición del grado de rotación de la rodilla, la laxitud rotacional se ha medido con imágenes de resonancia magnética empleando un aparato especial que aplica una par de rotación a la rodilla (el Porto-Knee Testing Device).

En los últimos años se ha desarrollado una serie de dispositivos mecánicos como, por ejemplo, el Robotic Knee Testing KSSTA, que consiste en unos servos y unas sujeciones en cadera y tobillo que permiten aislar la rotación del fémur con respecto a la tibia, se aplica un par al pie y un sensor electromagnético aplicado a la tibia mide la rotación.

Otros dispositivos mecánicos desarrollados por diversos autores son referidos por Colombet y col. (2012) destacando que presentan posibles movimientos entre el aparato y la pierna que afectan la medición del test.

La naturaleza estática de las restricciones al movimiento y, por tanto, del ensayo, además de la dificultad en la determinación de la posición de rotación neutra y también que hay que hacer gran cantidad de ensayos para abarcar todo el rango de movimiento, han provocado que los dispositivos para diagnosis tengan un desempeño deficiente.

Se concluye que a diferencia de los dispositivos de rehabilitación para terapia, los utilizados para el diagnóstico de lesiones se encuentran en una fase de menor desarrollo, pues en principio son muy pocos los desarrollos existentes y además, estos son evaluados negativamente por otros autores.

## GUÍA PARA EL DISEÑO DE DISPOSITIVOS PARA LA REHABILITACIÓN ROBÓTICA

Una constante en los artículos descritos hasta ahora es la escasa comparación con los proyectos previos y la información tan diversa en cuanto a los parámetros tomados en cuenta para diseñar cada dispositivo. En esta sección se pretende hacer una guía para el diseño de dispositivos de rehabilitación y diagnosis de extremidades inferiores basada en las consideraciones sobre otros artículos y en las recomendaciones propias de los autores.

Esta guía para el diseño de dispositivos de rehabilitación y diagnosis de extremidades inferiores está estructurada como una serie de pasos a seguir en la que en cada uno de ellas se presentan datos y recomendaciones.

## ANÁLISIS DE LA MOVILIDAD DE LA ARTICULACIÓN

Lo primero que se debe tomar en cuenta en el diseño de un dispositivo de rehabilitación y/o diagnosis de extremidades inferiores es el análisis de la movilidad de la articulación o las articulaciones que se desean estudiar.

En este caso, la articulación de la rodilla no es un simple un sistema de rotación de un grado de libertad, como pareciera ser a simple vista. Existen movimientos ampliamente estudiados que ocurren cuando la rodilla se flexiona durante la caminata, entre otros, hay una rotación de la pierna alrededor del eje longitudinal que puede ocurrir normalmente en personas sanas cuando están caminando.

De hecho, se afirma que la articulación de la rodilla constituye uno de los elementos más complejos del cuerpo humano debido a su diseño (García y col., 2003). Según estos autores, la articulación tiene un solo grado de libertad, la flexión-extensión, pero de manera accesoria posee una segunda rotación sobre el eje longitudinal de la pierna, la cual solo aparece cuando la rodilla está flexionada.

Además de conocer la movilidad de la articulación, es necesario tomar en cuenta algunas consideraciones relacionadas con la articulación, por ejemplo, si el centro de rotación del efector final de cualquier mecanismo se superpone sobre el centro de rotación de la articulación, ese dispositivo es capaz de permitir un correcto uso de las articulaciones y músculos (Malosio et al, 2012).

Finalmente, es importante aclarar que los modelos actuales que simulan la articulación de la rodilla no son exactos, sino que se aproximan al movimiento real de la articulación (Sanjuán y col., 2005), son conocidos al menos seis modelos que simulan la articulación de la rodilla. De hecho, la articulación de la rodilla y, en general, la mayor parte de las articulaciones humanas, tienen una configuración muy complicada.

## DEFINICIÓN DE LA ETAPA DE REHABILITACIÓN

Al tiempo que se estudia la articulación a rehabilitar o diagnosticar se puede definir en qué etapa de la rehabilitación se va a ubicar el mecanismo, mientras que para la diagnosis se trata de definir qué tipo de prueba que se quiere para el mecanismo.

Por lo general, la terapia postoperatoria consta de tres fases fundamentales, la primera busca devolver el movimiento natural e indoloro de la articulación, la segunda recupera la estabilidad y fuerza natural en la articulación, y la tercera recupera la función propioceptiva de la articulación. Normalmente, se aplican tratamientos termofísicos junto con la rehabilitación mecánica (López y col., 2013). Por su parte, la mayoría de los sistemas robóticos terapéuticos desarrollados ejecutan solo un tipo de ejercicios, bien sea pasivo, asistivo o resistivo (Akdogan y Adli, 2011).

Durante la primera fase de la rehabilitación se hacen ejercicios isométricos y pasivos, un proceso de apenas una semana de duración. Luego, comienzan los ejercicios activos que mejoran el rango de movimiento, y alrededor de la cuarta semana se logra una extensión completa y una flexión a $90^0$, comenzando también la recuperación de la propiocepción. Luego de un mes agregando los ejercicios para lograr la potenciación muscular, se alcanza la hiperextensión y flexión completa. También se comienza con los ejercicios de cadena cinemática cerrada y de cadena cinemática abierta. Ya en este momento, aproximadamente luego del segundo mes de la rehabilitación, los ejercicios comienzan a tomar un mayor nivel de exigencia y, en consecuencia, mayor independencia por parte del paciente, y se siguen haciendo ejercicios de propiocepción.

Posteriormente, el paciente hace ejercicios como carreras o natación, entre otros (Ramos y col., 2008).

Se reporta también la pertinencia de los ejercicios de cadena cinemática cerrada por ser más seguros, pues aumentan la estabilidad de la articulación y protegen a la articulación de fuerzas transversales/cortantes de desplazamiento anterior, además de que reproducen mejor la biomecánica normal estimulando así la propiocepción y la funcionalidad (Ramos et al, 2008). Otros estudios afirman que se ha observado que hay más actividad muscular con movimientos combinados, lo cual acelera los tiempos de recuperación en la terapia (Valdivia y col., 2013).

Se deduce entonces de lo anterior que durante una gran parte del tratamiento se efectúan ejercicios del tipo activo, mientras que los ejercicios pasivos se hacen en un tiempo mucho menor. Los ejercicios de propiocepción se ejecutan durante la mayor parte del tratamiento y finalmente, para la fase final del tratamiento, los ejercicios son implementados sin la necesidad de mecanismo, pues son ejecutados principalmente en espacios abiertos, tales como caminatas, carreras y natación, entre otros.

En lo relativo a los tratamientos para la diagnosis de lesiones es destacable el hecho de que gran parte de las lesiones ocurridas en las rodillas son del ligamento cruzado anterior, lo que hace que sea muy buena opción las pruebas que evalúen el daño en este ligamento. De todos los tratamientos para diagnosis de lesiones en la rodilla, el test de Lachman es quizá el más popular de todos porque tiene una sensibilidad del 87 al 98%, considerándose la prueba clínica de elección para el diagnóstico. Con menor sensibilidad diagnóstica disponemos de las maniobras del cajón anterior y del pivote (Ramos y col., 2008).

## MOVILIDAD DEL MECANISMO

A pesar de que la articulación tenga una cantidad específica de movimientos, es fundamental definir cuántos de ellos van a ser necesarios dependiendo del tipo de ejercicios o tratamientos que requiera la rehabilitación y diagnosis adoptada.

Algunos autores han definido esta fase del diseño como una de las más importantes. Afirman Saglia y col. (2010) que los protocolos de rehabilitación son considerados la base para el diseño de las estrategias de control. El mecanismo debe tratar de acoplarse adecuadamente a los movimientos de las articulaciones para evitar problemas de propiocepción en el paciente (Araujo-Gómez et al, 2016).

De esta manera, puede que su articulación presente una variedad específica de movimientos; sin embargo, la movilidad que presente el mecanismo quedará definida por el tipo de ejercicios que sea capaz de hacer. También se reporta que para que el cerebro pueda cambiar un patrón motor es necesaria la aplicación de cientos de repeticiones (Cioi y col., 2011), lo cual resulta ser siempre muy fácil de lograr con cualquier mecanismo que se diseñe.

En definitiva, es realmente importante definir qué movimientos cumplirá el mecanismo en concordancia con los protocolos de rehabilitación a los que se oriente.

## DISEÑO DEL MECANISMO

Esta etapa es obviamente la más importante, sin embargo tenemos ya cierto número de aspectos definidos que van a limitar el diseño, pues ya se debe saber entre otras cosas el tipo de ejercicios que debe cumplir el mecanismo.

Desde el punto de vista mecánico, la movilidad del mecanismo es la parte más importante del diseño, pero hay una serie de elementos que se deben considerar para mejorar su efectividad, desde unos tan relevantes como la ergonomía hasta otros que en principio no parecen ser tan obvios, como la confianza o la fe del paciente en el dispositivo. Se ha afirmado incluso que el diseño conceptual está relacionado con requerimientos del paciente, ya que las demandas del usuario resultan dominantes en el diseño conceptual (Yang, Zhu y Yao, 2009).

Entre los aspectos más importantes, muchos autores coinciden en que hay que considerar las exigencias de los pacientes para diseñar el mecanismo, y que la existencia de videojuegos o de ambientes virtuales que apoyen el funcionamiento del este son relevantes, especialmente cuando los pacientes son niños (Araujo-Gómez y col., 2016). Cioi y col. (2011) aseveran que los juegos de videos acoplados al proceso de rehabilitación motivan a los niños hasta tal punto que hacen repeticiones bajo cargas para superar con éxito el juego.

Por otra parte, una tendencia actual en muchos campos que involucran la interacción con humanos está en permitir que las actividades que estos llevan a cabo puedan ser llevadas a sus casas en virtud de generar ahorros principalmente relacionados con el transporte, por cuya razón es importante que los dispositivos sean lo más compactos posibles para que puedan ser transportados hasta distintos sitios. Adicionalmente, esto exigiría la posibilidad de que el terapeuta pueda monitorear la ejecución de los ejercicios a distancia (Girone y col., 2000).

Otro aspecto destacable se relaciona con la comodidad del paciente, lo que incide también en la aceleración del proceso de recuperación al sentirse a gusto el paciente ejecutando los ejercicios en el dispositivo. Girone y col. (2000) proponen mejoras en las correas de fijación del pie, una silla ajustable y un amortiguador para estabilizar la rodilla con el fin de mejorar su dispositivo.

También son importantes aspectos como la versatilidad del dispositivo para hacer una amplia variedad de ejercicios, así como su adaptabilidad a distintos pacientes en cuanto a tamaño y peso se refiere (Alcocer y col., 2012).

En la economía y simplicidad del mecanismo hay un consenso global entre todas las investigaciones en el área, y muy frecuentemente se valora la necesidad de que los dispositivos puedan presentar la mayor cantidad de data sobre la evolución del paciente en el proceso de rehabilitación. Algunos estudios también se enfocan en la integración de información electromiográfica para el control del dispositivo (Saglia y col., 2010).

En cuanto a su funcionamiento mecánico, una de las consideraciones que poco se han tomado en cuenta en el diseño del mecanismo es la resistencia que ofrece la pierna durante la ejecución del ejercicio, más importante aún en mecanismos diseñados para hacer ejercicios activos durante los cuales el paciente hace uso de fuerzas importantes durante un ejercicio. Se considera que para mecanismos que puedan ser ajustados por manipulación del terapeuta, es decir, cuando este pueda planificar una trayectoria, el mecanismo debería ofrecer la oposición que genera la pierna humana cuando el terapeuta aplica el tratamiento. En definitiva, resulta conveniente considerar la extremidad humana como un elemento más del mecanismo, lo cual puede equivaler a una cadena cinemática adicional.

En mecanismos de rehabilitación de extremidades inferiores es preferible utilizar los paralelos en busca de mejorar la capacidad de carga debido a la aplicación de fuerzas que, generalmente, son grandes (Satici y col. 2009), y se aumenta la capacidad de carga aún más si los actuadores son ubicados en la base de cada brazo (Pond y Carretero, 2004). También, la existencia de un brazo central en un mecanismo paralelo permite la reducción de las fuerzas de los motores comparado con mecanismos paralelos que no poseen el brazo central (Patanè y Cappa, 2011).

Una de las desventajas que presentan los dispositivos de rehabilitación para extremidades inferiores es que, generalmente, el centro instantáneo de rotación se encuentra por debajo de la pierna del paciente, lo que puede provocar que este experimente una propiocepción psicológica no natural (Malosio y col., 2012).

Jamwal y col. (2009) manifiestan que cuando se utilizan mecanismos paralelos para rehabilitación de extremidades inferiores, la existencia de plataformas en los dispositivos produce traslaciones y rotaciones que causan cambios en las piernas del paciente, por lo que el modelo dinámico debe incluir la inercia de la pierna. Además, en el momento en que el pie y la plataforma comienzan a moverse se genera cierta inseguridad en la posición del pie que conlleva a dificultades en el sistema de control.

También en esta sección se considera lo referido a los pares que se utilizarán en el dispositivo. Se ha reportado que el uso de pares prismáticos en los dispositivos de rehabilitación presenta el problema de que no responden muy bien a movimientos reversibles, mientras que los pares de revolución tienen una mejor respuesta (Liu y col., 2006; Alcocer y col, 2012), aunque, también, si queremos mejorar la ejecución de movimientos en reversa, reducir la fricción y la inercia al mínimo es una muy buena opción (Wheeler, Krebs y Hogan, 2004).

Adicionalmente, la posición de los actuadores, especialmente en dispositivos exoesqueléticos, produce fuerzas internas en las articulaciones que no se corresponden con los movimientos naturales de las extremidades y podrían producir un efecto errático en la coordinación motora del paciente (Malosio y col., 2012).

EVALUACIÓN DEL MECANISMO

Definido ya el diseño del mecanismo se procede a evaluar algunos aspectos cinemáticos como el espacio de trabajo, la versatilidad para hacer los ejercicios, la capacidad de repetirlos o de retornar rápidamente una posición, así como la maniobrabilidad o destreza, para estudiar si el mecanismo diseñado es capaz de cumplir los ejercicios de rehabilitación y diagnosis, así como también para redimensionar el mecanismo inicial.

Existen parámetros muy importantes para evaluar el mecanismo, entre otros, el número de condición. Según Jamwal y col. (2009) para obtener una mejor la controlabilidad del robot, el número de condición debe estar cerca de la unidad y, por tanto, requiere ser minimizado.

Por su parte, el espacio de trabajo es otro de los parámetros que se deberían evaluar siempre, en principio para determinar si el mecanismo es capaz de cumplir las tareas que se proponen y también porque ayuda a redefinir las dimensiones iniciales del mecanismo e incluso la selección de los pares y actuadores. En cuanto a las juntas utilizadas se afirma que las fuerzas aplicadas a los actuadores deben ser minimizadas para mejorar el funcionamiento del robot (Jamwal y col., 2009).

En la literatura se encuentran distintos parámetros de cada mecanismo que son evaluados, tales como capacidad de carga del mecanismo, repetitividad, costos del dispositivo y singularidades presentes. De todos estos se considerará que la singularidad es uno de los aspectos clave en la evaluación del mecanismo, pues la presencia de puntos de singularidad en él puede afectar seriamente la ejecución de los ejercicios por parte del paciente.

COMPARACIÓN CON OTROS MECANISMOS

Cada autor da sustento a su investigación, pero rara vez presenta una comparación con mecanismos desarrollados por otros autores o por sí mismo. En la actualidad hay muchos modelos de rehabilitación de extremidades inferiores, por lo cual sería interesante hacer una comparación entre algunos de ellos, así como ver la semejanza que hay entre los ejercicios cumplidos por el dispositivo y los ejecutados por un terapeuta.

Con este último paso se concluye la guía propuesta para el diseño del dispositivo de rehabilitación. Así pues, en la siguiente sección se presenta el diseño de un dispositivo en particular y las consideraciones tomadas en cuenta para cada una de los pasos que se han definido.

## CASO DE ESTUDIO: ROBOT PARALELO RPU+3UPS

En esta sección se presenta el diseño conceptual de un mecanismo para la rehabilitación de extremidades inferiores que sigue los pasos definidos en el apartado anterior.

## ANÁLISIS DE LA MOVILIDAD DE LA ARTICULACIÓN

Los movimientos relevantes en una rodilla son dos traslaciones y dos rotaciones. Las rotaciones ocurren en el eje transversal (eje laterolateral) con la flexión y extensión, que es el movimiento más importante de la rodilla, y la otra rotación se produce sobre el eje longitudinal de la pierna (eje craneocaucal), que es de menor grado, generando una rotación interna y externa de la rodilla. Aunque la tercera rotación existe, se puede considerar irrelevante, pues se produce en un ángulo muy pequeño y normalmente no es considerada para la aplicación de ejercicios de rehabilitación ni de diagnosis.

Las traslaciones de la articulación de la rodilla son bastante pequeñas, de hecho, las dos traslaciones en la dirección del eje vertical y en la dirección del eje anteroposterior se limitan a algunos centímetros. Sin embargo, el movimiento del pie durante la caminata presenta desplazamientos importantes en estas dos direcciones, que es donde se situará el enlace entre la pierna y el mecanismo. El desplazamiento en el eje transversal se considera insignificante en muchos tratamientos de diagnosis y de rehabilitación.

En consecuencia, vemos que los movimientos relevantes de la articulación de la rodilla son dos traslaciones y dos rotaciones, y de esta manera se limita la búsqueda de dispositivos del tipo 2T2R (dos traslaciones y dos rotaciones) con un eje común para una rotación y una traslación.

Se puede definir el modelo de articulación de la rodilla que se va a utilizar, sin embargo, este ha sido un tema ampliamente estudiado y se encuentra ampliamente documentado. En este texto se obviará la definición del modelo de la articulación de la rodilla.

## DEFINICIÓN DE LA ETAPA DE REHABILITACIÓN

El mecanismo a diseñar se orientará a la fase intermedia de la rehabilitación debido a que en ella se exige mayor movilidad que en la parte inicial y además la articulación está comenzando a tomar fuerza, por lo que los ejercicios que debe hacer el paciente son de tipo pasivo y activo. Es esta la fase que requiere mayor esfuerzo por parte del fisioterapeuta.

En la primera fase de rehabilitación, en la cual se aplican ejercicios en su mayoría pasivos, se han desarrollado dispositivos que reproducen un movimiento continuo pasivo (CPM). Hay algunos estudios que demuestran que los sistemas de movilidad pasiva continua garantizan un mayor beneficio, sin embargo, el elevado costo de estos no justifica su utilización (Ramos y col, 2008).

En lo que a la diagnosis se refiere, el test de Lachman y el desplazamiento de pivote son dos de los más reconocidos y aplicados en la actualidad, principalmente cuando existen lesiones de ligamento cruzado anterior, se considera apropiado que el dispositivo sea capaz de cumplir estas pruebas.

## MOVILIDAD DEL MECANISMO

La articulación de la rodilla puede rotar alrededor de los tres ejes ortogonales, aunque en términos de requerimientos de rehabilitación, en la rodilla es importante la rotación alrededor del eje transversal y del eje vertical. Además, gran cantidad de ejercicios de rehabilitación ocurren con el desplazamiento de la rodilla en el plano sagital (Araujo-Gómez y col., 2016).

Ejercicios para rehabilitación como, por ejemplo, simular el movimiento de la caminata, exigen a la rodilla el trasladarse en el plano sagital y rotar alrededor del eje transversal.

El test de Lachman se efectúa con desplazamientos en el plano sagital, mientras que de desplazamiento del pivote exige a la articulación que rote en el eje vertical. En conclusión, los ejercicios de rehabilitación y diagnosis que el dispositivo cumplirá ocurren con los desplazamientos y rotaciones indicadas, por lo que un robot con 4 GDL, 2T2R cubriría los ejercicios propuestos.

DISEÑO DEL MECANISMO

En primer lugar, debido a la aplicación de este mecanismo se entiende que el dispositivo debe soportar fuerzas importantes, pues sobre la rodilla, y más aún sobre el pie, puede aplicarse un porcentaje importante del peso total del cuerpo humano. De esta manera se decide implementar un mecanismo paralelo de 4 GDL por la alta capacidad de carga que tienen estos mecanismos, además de considerar conveniente que uno de los brazos del mecanismo se ubique justo en su centro, lo que aportaría estabilidad y capacidad de carga al mecanismo.

Otro parámetro que delimita el diseño del mecanismo está relacionado con su movimiento. En este sentido, según lo observado hasta el momento, el mecanismo no debe moverse en una dirección específica (definida como y), por lo que el desplazamiento se mantiene limitado en un plano (plano xz). Adicionalmente, en uno de los ejes que quedan contenidos en este plano (se selecciona x) no debe existir rotación.

También resulta conveniente que el mecanismo pueda cumplir ejercicios de cadena cinemática cerrada, lo que implica en líneas generales que el pie debe estar ajustado firmemente a él.

Se deben considerar parámetros para el mecanismo que mejoren su funcionabilidad, entre otros, un aspecto agradable para el paciente y capacidad de almacenar datos sobre la evolución del paciente.

Volviendo al análisis de la movilidad requerida por el mecanismo, este movimiento requerido se puede lograr de distintas maneras. Una posibilidad es que el brazo central se mueva en un mismo plano a partir de un par de revolución y un par prismático, así, el brazo central, desde el punto en donde se conecta a la plataforma fija hasta el punto en donde se conecta a la plataforma móvil, permanece siempre en un mismo plano.

Por otro lado, según Merlet (2012), para que un mecanismo paralelo de 4 GDL pueda tener el movimiento propuesto, es imposible que los cuatro brazos de que dispone el mecanismo sean idénticos, por lo que se analiza la posibilidad de que los tres brazos adicionales al brazo central sean similares entre sí pero distintos al central.

Para el brazo central ya se definió que debe tener un par de revolución y un par prismático. Para completar los 4 GDL se puede agregar una junta universal, los cual implica que el brazo central es de configuración RPS o semejante. Los brazos laterales deben sumar 6 GDL a partir de tres juntas, y una posible solución es UPS o algo semejante.

Entre las posibles configuraciones se han analizado las siguientes: una primera configuración del tipo RPU+3UPS y otra del tipo PRU+3PUS (ver Fig. 8).

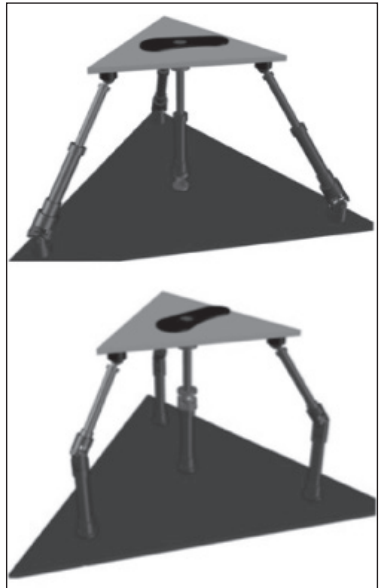

Figura 8. Propuesta RPS+3UPS y PRS+3PUS

## EVALUACIÓN DEL MECANISMO

Entre las distintas alternativas para evaluar la factibilidad del mecanismo está el verificar si el mecanismo es capaz de desplazarse cuando el paciente esté haciendo un ejercicio que simule el movimiento de la caminata, de esta manera se debe hacer un análisis cinemático en el que se determine el espacio de trabajo del mecanismo y verificar si dentro de ese espacio se puede ubicar el movimiento de la pierna durante la caminata.

Para definir el espacio de trabajo se parte de resolver el problema cinemático inverso, por lo cual se determina el tamaño de los pares prismáticos necesarios para definir una posición específica del mecanismo. El diseño mecatrónico, incluyendo la cinemática del mecanismo, puede ser consultada en Araujo y col., (2017).

## COMPARACIÓN CON OTROS MECANISMOS

Al igual que en la fase anterior, nos limitaremos a evaluar solamente un aspecto de la cinemática del mecanismo. En este caso se comparan los espacios de trabajo que tienen ambos mecanismos propuestos en este capítulo para evaluar la conveniencia de la elección efectuada.

La figura 9 muestra el espacio de trabajo que cubre cada mecanismo propuesto. Se representa el desplazamiento de los dos modelos en el plano *xz* cuando las rotaciones alrededor de los ejes *y* y *z* son nulas.

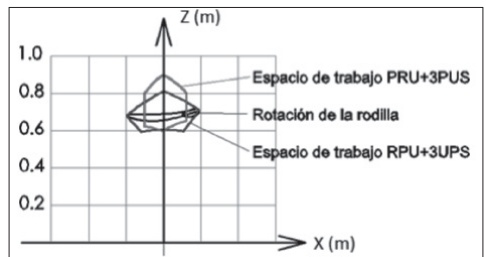

Figura 9. Comparación de los espacios de trabajo de los dos modelos (PRU+3PUS, RPU+3UPS)

Aunque es un solo aspecto el que se está evaluando, se pueden extraer algunas conclusiones como, por ejemplo, el hecho de que el modelo PRU+3PUS tiene mayor movilidad en una dirección vertical mientras que el RPU+3UPS presenta mayor movilidad en una dirección horizontal, lo que lo hace más adecuado para el ejercicio de la caminata.

CONCLUSIONES

Este capítulo contempla una guía para el diseño de mecanismo de rehabilitación de extremidades inferiores, aunque más que una serie de pasos a seguir, lo que busca es recopilar muchas recomendaciones que se encuentran en la literatura existente en el área, además de las recomendaciones del autor durante el diseño y construcción de este tipo de dispositivos. Adicionalmente se pretende que esta guía sirva de referencia para el diseño de dispositivos de rehabilitación en otras partes del cuerpo humano.

En la parte final del capítulo se presenta un caso de estudio en el que la guía propuesta es empleada para el desarrollo de un robot paralelo de 4-GdL para la rehabilitación de extremidad inferior. El caso de estudio finaliza con una comparación entre dos dispositivos propuestos por los autores. Futuros trabajos se recomiendan comparar distintos dispositivos presentados en la literatura. De hecho, sería muy valioso hacer un estudio que compare los dispositivos más conocidos y recomendados en la actualidad

REFERENCIAS